\pdfoutput=1

\documentclass[11pt]{article}

\usepackage{acl}

\usepackage{times}
\usepackage{latexsym}
\usepackage{amsmath}
\usepackage{amssymb}
\usepackage[capitalise]{cleveref}
\usepackage[T1]{fontenc}

\usepackage[utf8]{inputenc}

\usepackage{microtype}

\usepackage{graphicx}
\usepackage{pdfpages}
\usepackage{booktabs}

\newcommand{\Single}{{\scshape Single}}
\newcommand{\Multi}{{\scshape Multi}}
\newcommand{\Any}{{\scshape Any}}

\newcommand{\dataset}{ViComTe} 
\makeatletter
\def\thanks#1{\protected@xdef\@thanks{\@thanks
        \protect\footnotetext{#1}}}
\makeatother

\author{Chenyu Zhang \qquad 
        Benjamin Van Durme \qquad
        Zhuowan Li\footnotemark[1]
        \thanks{\llap{\textsuperscript{*}}{Joint Advising}} 
        \qquad 
        Elias Stengel-Eskin\footnotemark[1] \\
  Johns Hopkins University \\
\texttt{\{czhan105, vandurme,  zli110, elias\}@jhu.edu}}

\title{Visual Commonsense in Pretrained Unimodal and Multimodal Models}

\begin{document}
\maketitle
\begin{abstract}
Our commonsense knowledge about objects includes their typical visual attributes; we know that bananas are typically yellow or green, and not purple.
Text and image corpora, being subject to reporting bias, represent this world-knowledge to varying degrees of faithfulness.  
In this paper, we investigate to what degree unimodal (language-only) and multimodal (image and language) models capture a broad range of visually salient attributes. 
To that end, we create the 
Visual Commonsense Tests (\dataset) dataset
covering 5 property types (color, shape, material, size, and visual co-occurrence) for over 5000 subjects.
We validate this dataset by showing that our grounded color data correlates much better than ungrounded text-only data with crowdsourced color judgments provided by \citet{paik-etal-2021-world}. 
We then use our dataset to evaluate pretrained unimodal models and multimodal models. 
Our results indicate that multimodal models better reconstruct attribute distributions, but are still subject to reporting bias. 
Moreover, increasing model size does not enhance performance, suggesting that the key to visual commonsense lies in the data.\footnote{The dataset and code is available at {\url{https://github.com/ChenyuHeidiZhang/VL-commonsense}}.}

\end{abstract}

\section{Introduction}

\begin{figure*}[ht]
    \centering
    \includegraphics[width=\textwidth-30pt]{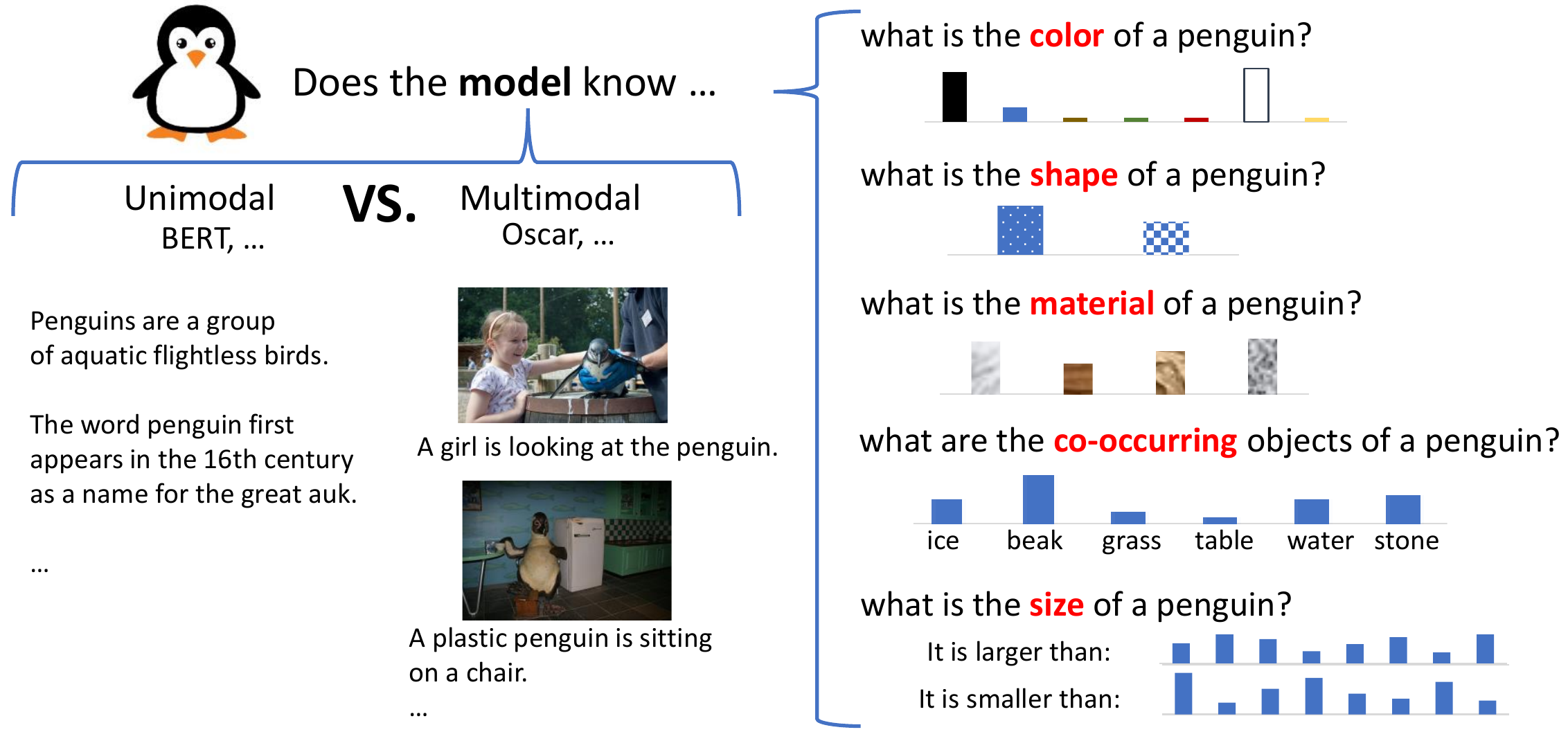}
    \caption{
    We compare unimodal and multimodal models' abilities to capture visual commonsense knowledge. The commonsense knowledge is evaluated on five relation types: color, shape, material, size, and visual co-occurrence. We compare the model outputs with the gold distribution from \dataset, which is mined from Visual Genome. }
    \label{fig:fig1}
\vspace{-0.5em}
\end{figure*}


The observation that human language understanding happens in a rich multimodal environment has led to an increased focus on visual grounding in natural language processing (NLP)  \citep{multimodal-ml,bisk-etal-2020-experience}, driving comparisons between traditional unimodal text-only models and multimodal models which take both text and image inputs. 
In this work, we explore to what extent unimodal and multimodal models are able to capture commonsense visual concepts across five types of relations: color, shape, material, size, and visual co-occurrence (cf. \cref{fig:fig1}).
We further explore how this ability is influenced by reporting bias \citep{Gordon2013ReportingBA}, the tendency of large corpora to over- or under-report events.
We define visual commonsense as knowledge about generic visual concepts, e.g. "knobs are usually round," and we measure this knowledge via frequency distributions over potential properties (e.g. round, square, etc).
A visually-informed language model should be able to capture such properties. Our color, shape, material, and co-occurrence data are mined from Visual Genome \citep{VG}, and our size data are created from object lists. 
They contain a large number of examples of per-object attribute distributions and ``object-attribute'' pairs.

\citet{paik-etal-2021-world} evaluate language models' color perception using a human-annotated color dataset (CoDa), finding that reporting bias negatively influences model performance and that multimodal training can mitigate those effects. 
In this work, we confirm those findings while extending the evaluation to a broader range of visually salient properties, resulting in a more comprehensive metric for visual commonsense. 
In order to elicit visual commonsense from language models, we utilize soft prompt tuning \citep{soft-prompt}, which trains optimal templates by gradient descent for each model and relation type that we explore. 
We also utilize knowledge distillation to enhance a text-only model's visual commonsense ability, where the vision-language model serves as the teacher.

The major contributions of this work are: (1) we design a comprehensive analytic dataset, \dataset, for probing English visual commonsense, that is applicable to any language model; (2) we use {\dataset} to study models' ability to capture empirical distributions of visually salient properties. We examine unimodal language models, multimodal vision-language (VL) models, and a knowledge-distilled version of a VL model; and (3) we analyze the effects of reporting bias on the visually-grounded vs. ungrounded datasets and models.%

\section{Related Work}

\subsection{Vision-Language Modeling}

Recent advances in vision-language (VL) modeling have led to increased success on benchmark tasks. Most VL models learn joint image and text representations from cross-modal training of transformers with self-attention, including LXMERT \citep{tan2019lxmert}, ViLBERT \citep{Lu2019ViLBERT}, VisualBERT \citep{li2019visualbert}, UNITER \citep{uniter}, etc.
Oscar \citep{li2020oscar} additionally uses object tags in images as anchor points to facilitate the learning of image-text alignments and VinVL \citep{zhang2021vinvl} presents an improved object detection model. CLIP \citep{clip} learns by predicting caption-image alignment from a large internet corpus of (image, text) pairs.

While our work uses textual prompt tuning techniques, there have also been work on visual prompt engineering to enhance the performance of pretrained vision-language models. \citet{zhou2021coop} model context in prompts as continuous representations and learn to optimize that context.
\citet{Yao2021CPTCP} develop a cross-modal prompt tuning framework that reformulates visual grounding as a fill-in-the-blank problem for both image and text.

\subsection{Visual Commonsense}

In one of the early attempts at learning visual commonsense, \citet{Vedantam2015LearningCS} measure the plausibility of a commonsense assertion in the form of (obj1, relation, obj2) based on its similarity to known plausible assertions, using both visual scenes and accompanying text. \citet{zellers-etal-2021-piglet} learn physical commonsense via interaction, and use this knowledge to ground language. \citet{Frank2021VisionandLanguageOV} probe whether VL models have learned to construct cross-modal representations from both modalities via cross-modal input ablation. 

Note that our definition of visual commonsense differs from that of  \citet{zellers2019vcr}, where the model is required to perform commonsense reasoning based on an image.
Our definition of visual commonsense is more similar to the idea of stereotypic tacit assumptions \citep{prince1978} -- the propositional beliefs that humans hold about generic concepts, such as ``dogs have to be walked.'' \citet{tacit-assumption} probe neural language models for such human tacit assumptions and demonstrate the models' success. We extend this intuition to visual concepts and explore how visual information may help language models to capture such assumptions.

There has also been earlier work on the McRae feature norms \citep{McRae2005SemanticFP}, in which human annotators wrote down attributes that describe the meaning of words. For instance, ``car'' can be labeled as ``has four wheels'' and ``apple'' can be labeled as ``is green.''
\citet{Silberer2013ModelsOS} expand the McRae dataset into a set of images and their visual attributes and construct visually grounded distributional models that can represent image features with visual attributes.

\citet{language-prior} examine the ``language prior'' problem in Visual Question Answering models, where models tend to answer based on word frequencies in the data, ignoring the image contents. In this work, we explore to what extent such a language prior is recruited absent a visual input.

\subsection{Reporting Bias}
\label{reporting-bias}

Pretrained language models such as BERT \cite{devlin-etal-2019-bert} are trained on billions of tokens of text, capturing statistical regularities present in the training corpora.
However, their textual training data can suffer from reporting bias, where the frequency distribution of specific events and properties in text may not reflect the real-world distribution of such properties \citep{Gordon2013ReportingBA}. For example, while grass is typically green, this may be under-reported in web corpora  (as it is assumed to be true), and while motorcycle crashes may be more common in the real world, plane crashes are mentioned far more in news text \citep{Gordon2013ReportingBA}.
\citet{misra2016seeing} highlight the reporting bias in ``human-centric'' image annotations and find that the noise in annotations exhibits a structure that can be modeled.

\section{Dataset: \dataset}

\subsection{Dataset Mining}

\begin{table*}
\small
\centering
\begin{tabular}{lrrll}
\hline
\textbf{Relation} & \textbf{\# Classes} & \textbf{\# (subj, obj) Pairs} & \textbf{Ex Template} & \textbf{Ex (subj, obj) Pair} \\
\hline
color & 12 & 2877 & [subj] {\it can be of color} [obj] & ({\it sky}, {\it blue}) \\
shape & 12 & 706 & [subj] {\it has shape} [obj] . & ({\it egg}, {\it oval}) \\
material & 18 & 1423 & [subj] {\it is made of} [obj] . & ({\it sofa}, {\it cloth}) \\
size (smaller) & 107 & 2000 & [subj] {\it is smaller than} [obj] . & ({\it book}, {\it elephant}) \\
size (larger) & 107 & 2000 & [subj] {\it is larger than} [obj] . & ({\it face}, {\it spoon}) \\
co-occurrence & 5939 & 2108 & [subj] {\it co-occurs with} [obj] . & ({\it fence}, {\it horse}) \\
\hline
\end{tabular}
\caption{\label{data-summary}
Summary of the {\dataset} dataset and the manual templates, including the number of classes, (subject, object) pairs, and an example pair for each relation.}

\hspace{1.4cm}

\begin{tabular}{llrrrrrr}
\hline 
\textbf{Source} & \textbf{Group} & \textbf{Spearman $\rho$} & \textbf{\# Subjs} & \textbf{Avg \# Occ} & \textbf{Top5 \# Occ} & \textbf{Btm5 \# Occ} & \textbf{Acc@1} \\ 
\hline
VG & All & 64.3 $\pm$ 23.9 & 355 & 1252.6 & 64.6 & 308.6\\
 & \Single{} & 62.2 $\pm$ 24.0 & 131 & 494.9 & 64.6 & 1181.6 & 80.2\\
 & \Multi{} & 69.3 $\pm$ 20.7 & 136 & 1156.1 & 2062.2 & 347.0\\
 & \Any{} & 58.4 $\pm$ 27.1 & 88 & 2529.6 & 8452.4 & 1213.4\\
\hline
Wikipedia & All & 33.4 $\pm$ 30.6 & 302 & 543.6 & 1758.0 & 49.8\\
 & \Single{} & 29.6 $\pm$ 29.9 & 110 & 352.2 & 345.8 & 35.0 & 35.5\\
 & \Multi{} & 33.9 $\pm$ 30.9 & 119 & 500.8 & 1242.0 & 27.6\\
 & \Any{} & 38.2 $\pm$ 30.4 & 73 & 902.0 & 3000.2 & 161.2\\
\hline
\end{tabular}
\caption{\label{data-eval} Evaluation of {\dataset} (mined from VG) and Wikipedia-mined color datasets by comparing with the human-annotated dataset CoDa. Reported are the average Spearman correlation ($\times 100$), number of common subjects, average number of occurrences of the common subjects, average number of occurrences of subjects with top- and bottom-5 Spearman correlations, and the percentage of top-1 attributes being matched for the single group. {\dataset} has higher correlations with human annotations.}
\end{table*}

For each relation color, shape, material, size, and object co-occurrence, our data take the form of (subject, object) tuples extracted from object distributions per subject. The goal is to predict the object and its distribution from the subject and relation. \cref{data-summary} summarizes the number of classes and subject-object pairs for each relation.\footnote{See \cref{sec:list-objs} for more information on the object classes.}

\paragraph{Color, Shape, Material}
For color, shape, and material, the subject is a noun and the object is the color, shape, or material property of the noun, mined from attributes of Visual Genome (VG) \citep{VG}.\footnote{Licensed under CC-BY 4.0.} We manually create a list of single-word attributes for each relation, and only VG subjects that are matched with a specific attribute for more than a threshold number of times are recorded, in order to avoid noise in the dataset. The thresholds for color, material, and shape are 5, 2, and 1, respectively, chosen based on the availability of attributes of each relation in VG. VG attributes are filtered with the following steps: (1) attribute ``Y colored / made / shaped'' is treated as ``Y''; (2) select only the last word for compound attributes (e.g. treat ``forest green'' as ``green''); (3) similar attributes are merged into a main attribute class (e.g. ``maroon'' and ``crimson'' become ``red'').

The above procedure produces a distribution over the set of attributes for each subject noun. 
From that distribution, a (subject, object) data instance is generated for each subject where the object is the attribute that associates with it the most.
See the first three rows of \cref{data-summary} for examples.

\paragraph{Size}
Size is separated into {\tt{size\_smaller}} and {\tt{size\_larger}}, where the subject is a noun and the object is another noun that is smaller or larger, respectively, than the subject. 
To form the size dataset, we obtain a set of concrete nouns that appears in VG, which we manually classify into 5 size categories ({\tt{tiny}}, {\tt{small}}, {\tt{medium}}, {\tt{large}}, and {\tt{huge}}). Typical objects in each category includes \emph{pill}, \emph{book}, \emph{table}, \emph{lion}, \emph{mountain}, respectively.
We randomly pick two nouns from different categories to form a (subject, object) pair. 

\paragraph{Visual Co-occurrence}
The visual co-occurrence dataset is generated in a similar way to the color, shape, and material datasets. Co-occurrence distribution is extracted from Visual Genome where two objects that occur in the same scene graph together for more than 8 times are recorded, and a (subject, object) instance is generated for each subject, where the object is the noun that co-occurs with the subject the most.

\subsection{Data Grouping}
Following \citet{paik-etal-2021-world}, we split the color, shape, and material datasets each into three groups: \Single{}, \Multi{}, and \Any{}. The \Single{} group is for subjects whose most common attribute covers more than 80\% of the probability, e.g., the color of \emph{snow} is almost always white. The \Multi{} group is defined as subjects not in the \Single{} group where more than 90\% of the probability falls in the top 4 attribute classes, e.g., the color of a penguin in \cref{fig:fig1}. The rest of the subjects are in the \Any{} group. Lower model performance for the \Single{} group would indicate the influence of reporting bias. For example, if the model is unable to correctly capture the distribution of the color of \emph{snow}, it is likely because the color of snow has low probability of being reported in the training corpus, as people know it is white by default.

\subsection{Templates}
In order to elicit model response and extract target objects and distributions from text, we manually design a set of templates for each relation. There are 7 templates for color, shape, and material each, 8 for size, and 4 for visual co-occurrence.
See \cref{data-summary} for example templates.

\subsection{Wikipedia Data}
In order to compare text-based and visually-grounded data, we mine the color, shape, and material datasets from Wikipedia data, which is typically used in model pretraining. 
To mine these text-based datasets, we combine the sets of subjects in VG, take the manual list of attributes as objects again, and extract (subject, object) pairs if the pair matches any of the pre-defined templates. In \cref{sec:data_eval} we will show the advantages of the VG-mined dataset over this text-based dataset.

\subsection{Dataset Evaluation} \label{sec:data_eval} 

To ensure the validity of {\dataset}, we compare our color dataset with the human-annotated CoDa dataset \citep{paik-etal-2021-world}, which we assume is close to real-world color distributions and has minimal reporting bias.
We see a reasonably strong correlation with CoDa, indicating that the {\dataset} dataset is a good and cost-effective approximation to human annotations.

\paragraph{Metrics}
We report the Spearman's rank-order correlation between the two distributions in comparison, averaged across all subjects. The Spearman correlation is used instead of the Pearson correlation since for our purpose the rank of the object distributions is more important than the exact values, which may change due to data variability.
The top-1 accuracy (Acc@1) is the percentage of the objects with the highest probability in the source distributions matching those in the target distributions.
These two metrics are also used in later sections when evaluating model distributions.

\paragraph{Analysis}

\cref{data-eval} shows the detailed results of the evaluation of the {\dataset} and Wikipedia color datasets by comparing with the human-annotated dataset, CoDa. We can see that {\dataset} has much higher Spearman correlation with CoDa, as well as substantially higher top-1 accuracy for the \Single{} group. The correlation is expected to be low for the \Any{} group, because objects in the \Any{} group can have many possible colors.

Reporting bias is present in both datasets, as the average number of occurrences of \Single{} group subjects are much fewer than that of the \Multi{} and \Any{} group subjects. 
Counter-intuitively, for {\dataset}, the highly-correlated \Single{} group subjects have fewer average occurrences than the ones with low correlations. 
This is contrary to our expectation that more frequent objects would better reflect the human-perceived distribution
and can be explained by \Single{} subjects being easier to represent even without a large amount of data.

One example where the Wikipedia distribution diverges from the CoDa distribution is ``penguin,'' whose most likely color in CoDa is black, followed by white and gray; however, its top color in Wikipedia is blue, because ``blue penguin'' is a specific species with an entry in Wikipedia, even if it is not as common as black and white penguins. One example where the VG distributions diverge from CoDa is ``mouse,'' because in VG, most occurrences of ``mouse'' are computer mice, which are most commonly black, whereas when asked about the word ``mouse'', human annotators typically think about the animal, so that the most likely colors in CoDa are white and gray.\footnote{Additional examples are provided in \cref{sec:data-analysis}.}

\subsection{Dataset splits}
Each of the color, shape, material, size, and co-occurrence datasets is split into 80\% training data and 20\% test data. All evaluation metrics are reported on the test set. The training set is used for the logistic regression and the soft prompt tuning algorithm (\cref{sec:eval}).

\section{Probing Visual Commonsense}

\subsection{Models} \label{models}

We examine 7 pretrained transformer-based models and 2 variations of them, trained on a variety of data. BERT \citep{devlin-etal-2019-bert}, ALBERT \citep{lan2020albert}, and RoBERTa \citep{liu2019roberta} are trained on text only using a masked language modeling objective (MLM). 
Oscar \citep{li2020oscar} is a vision-language model based on the BERT architecture, trained with an combined MLM and contrastive loss on text-image pairs. 
VisualBERT \citep{li2019visualbert} is another vision-language model based on BERT that learns joint representation of images and text.
\citet{tan-bansal-2020-vokenization} introduce the ``vokenization'' method, which aligns language tokens to their related images, mitigating the shortcomings of models trained on visually-grounded datasets in text-only tasks. Since our task is purely text-based, we also experiment with a pretrained vokenization model (BERT + VLM on Wiki).
Finally, we use representations from CLIP (ViT-B/32) \citep{clip}, which is trained with a contrastive image-caption matching loss.

\paragraph{Distilled Oscar}
As our experiments involve exclusively textual inputs, we develop a knowledge-distilled version of Oscar (``Distilled'') which corrects for the lack of image input in our task.
Knowledge distillation \citep{hinton2015distilling, distil_bert} is the process of transferring knowledge from one model to another, where the student model is trained to produce the output of the teacher model. 
Here, we use Oscar as the teacher and BERT as the student.
The training data is part of the Oscar pretraining corpus: COCO \citep{coco}, Flickr30k \citep{flickr30k}, and GQA \citep{gqa}, and the Distilled Oscar model has access to the text data only.
We use the Kullback-Leibler loss to measure the divergence between the output logits of BERT and Oscar, and optimize the pretrained BERT on that loss to match the outputs of Oscar. Configurable parameters are set the same as for Oscar pretraining.

\paragraph{CaptionBERT}
Since VL models are trained largely on caption data, it could be that the differences between a text-only model and a VL model come not from a difference in modalities -- text vs. images and text -- but from a difference in domain -- webtext vs. image captions.
In order to disentangle the effects of the domain difference from those of visual inputs, we train a BERT model from scratch (``CaptionBERT'') on Oscar's caption-based text data (the same data as for the Distilled model). 
If CaptionBERT, which does not have exposure to visual inputs, performs better than BERT and similarly to VL models (which are trained with visual inputs), it would suggest that the training domain matters more than the modality.
If, on the other hand, CaptionBERT performs worse than VL models, it would highlight the importance of modality. 

\subsection{Evaluation Methods} \label{sec:eval}
We compare the visual commonsense abilities of pretrained unimodal and multimodal models.
Given a list of prompts and a subject word, each model outputs the distribution of the target word.
Following \citet{paik-etal-2021-world}, we apply zero-shot probes to models that are trained on a language modeling objective, and conduct representation probes for those that are not. We report the prediction accuracy and the Spearman correlation of the output distribution with the true distribution.

We use models trained with an MLM objective (BERT,  Distilled, etc) directly for zero-shot prediction of masked tokens.\footnote{For the target words that contain more than one subword tokens, we use the first token as the target.}
For Oscar we  add a word-prediction head on top of it.
The results across templates are aggregated in two modes.
In the ``best template'' mode, for each example, the highest Spearman correlation among all templates is reported, and the top-1 result is regarded as correct if the true target object is the same as the top-1 result of any of the templates.
In the ``average template'' mode, the output distribution is the mean of the distributions across all templates.

Since CLIP is not trained on a token-prediction objective, we implement logistic regression on top of the frozen encoder output, to predict the target attribute or object. The input is each of the templates with the subject [X] filled with an input in the dataset. Like \citet{paik-etal-2021-world}, to give the model ample chance of success, we take the template that results in the best test accuracy score, report that accuracy and the Spearman correlation associated with that template.
For the classification head, we use the Scikit-Learn implementation of Logistic Regression (random\_state=0, C=0.316, max\_iter=2000) \citep{scikit-learn}.

\paragraph{Soft prompt tuning}

In order to overcome the limitation of self-designed prompts, we incorporate prompt tuning technique that learns soft prompts by gradient descent, from \citet{soft-prompt}.\footnote{\url{https://github.com/hiaoxui/soft-prompts}} The algorithm minimizes the log loss:

$$\sum\limits_{(x,y) \in E_r} - \log \sum\limits_{\textbf{t} \in T_r} p(y | \textbf{t}, x)$$

\noindent for a set of example pairs $E_r$ and template set $T_r$.







\begin{table*}[ht]
\small
\centering
\begin{tabular}{cl|lr|lr|lr|l}
\hline
 & & \multicolumn{2}{c|}{\textbf{Color}} & \multicolumn{2}{c|}{\textbf{Shape}} & \multicolumn{2}{c|}{\textbf{Material}} & \textbf{Cooccur}\\
\cmidrule(lr){3-4}\cmidrule(lr){5-6}\cmidrule(lr){7-8}\cmidrule(lr){9-9}
\textbf{Tune} & \textbf{Model} & \textbf{Spearman $\rho$} & \textbf{Acc@1} & \textbf{Spearman $\rho$} & \textbf{Acc@1} & \textbf{Spearman $\rho$} & \textbf{Acc@1} & \textbf{Spearman $\rho$}\\
\hline
 & BERT$_b$ & 26.1 $\pm$ 31.0* & 11.7 & 38.7 $\pm$ 15.1 & 6.7 & 33.7 $\pm$ 19.6 & 30.0 & 4.7 $\pm$ 3.5\\
 & Oscar$_b$ & 26.4 $\pm$ 30.7* & 24.0 & 45.9 $\pm$ 14.1 & \textbf{53.0} & 38.6 $\pm$ 17.5 & \textbf{43.3} & 9.8 $\pm$ 6.9\\
 No & Distilled & 34.8 $\pm$ 27.3 & 27.5 & \textbf{46.2 $\pm$ 14.2} & 37.3 & 36.1 $\pm$ 20.2 & 37.7 & \textbf{10.1 $\pm$ 7.5}\\
 & BERT$_l$ & \textbf{37.6 $\pm$ 30.3} & \textbf{30.3} & 42.7 $\pm$ 17.1 & 28.4 & 36.6 $\pm$ 19.1 & 35.7 & 5.2 $\pm$ 3.8\\
 & Oscar$_l$ & 31.8 $\pm$ 28.3 & 17.1 & 40.0 $\pm$ 16.9 & 38.1 & \textbf{39.2 $\pm$ 17.1} & 40.5 & 9.7 $\pm$ 6.7\\
\hline
 & BERT$_b$ & 48.0 $\pm$ 22.9 & 47.4 & 49.2 $\pm$ 12.7* & 76.1 & 41.2 $\pm$ 15.3 & 45.2 & 11.3 $\pm$ 7.9\\
 & Oscar$_b$ & \textbf{58.1 $\pm$ 21.1} & \textbf{67.9} & 50.4 $\pm$ 11.5* & 81.3 & 45.3 $\pm$ 14.3 & \textbf{66.2} & 12.7 $\pm$ 9.3\\
Yes & Distilled & 57.1 $\pm$ 21.9 & 64.6 & \textbf{50.5 $\pm$ 12.3} & \textbf{82.8} & \textbf{45.4 $\pm$ 14.8} & \textbf{66.2} & \textbf{13.0 $\pm$ 10.1}\\
 & BERT$_l$ & 37.6 $\pm$ 30.3 & 30.3 & 49.2 $\pm$ 12.6 & 78.4 & 43.7 $\pm$ 15.1 & 53.3 & 11.4 $\pm$ 8.0\\
 & Oscar$_l$ & 57.6 $\pm$ 21.6 & 65.3 & 50.1 $\pm$ 12.2 & 81.3 & 45.2 $\pm$ 15.2 & 65.8 & 12.8 $\pm$ 9.6\\
\hline
\end{tabular}
\caption{\label{zero_shot_avg_all}
Spearman correlation and top-1 accuracy (both $\times$ 100) of zero shot probing, before and after soft prompt tuning (``N'' and ``Y'' for the ``Tune'' column). This is the ``average template'' case where the output distribution is the mean of distributions across all templates. The Spearman correlation reported is the mean across all subjects $\pm$ standard deviation, comparing the output distribution and the Visual Genome distribution. The subscripts $b$ and $l$ indicate the size of the model, and Distilled is the BERT model after distilling from Oscar. Asterisk indicates where there is \emph{no} significant difference between $\textnormal{BERT}_b$ and $\textnormal{Oscar}_b$ (t-test p-value > 0.05).}
\vspace{-1em}
\end{table*}



\subsection{Size Evaluation} \label{size-eval}
The size dataset differs from the other datasets in that we use relative sizes (X is larger/smaller than Y), as absolute size information is hard to obtain.
Thus, we use two evaluation strategies for size. 

\paragraph{Rank partition}
First, as in the previous prediction task, given a template such as ``[X] is larger than [Y]'' and an object [X], we ask the model to predict the distribution of [Y], taking only the distribution $D$ of nouns in the size dataset. For the current object [X], we take the nouns in size categories that are smaller than the category of [X] ($N_{sm}$), and those that are in larger categories ($N_{lg}$). Let the length of $N_{sm}$ be $m$ and the length of $N_{lg}$ be $n$. Then for the ``larger'' templates, we compute the average percentage of overlap between the top $n$ objects in $D$ and $N_{lg}$ and that between the bottom $m$ objects in $D$ and and $N_{sm}$. For the ``smaller'' templates, the ``top'' and ``bottom'' are reversed.

\paragraph{Adjective projection}
The second approach follows that of \citet{color-adj}, which projects the word to be evaluated onto an adjective scale. In this case, we compute the word embeddings of the adjectives ``small'' and ``large'' and the nouns from models, so the scale is $\overrightarrow{\textnormal{large}} - \overrightarrow{\textnormal{small}}$ and the projection is calculated by cosine similarity.
For instance, for the example noun ``bear'', the projection score is given by:
\vspace{-0.5em}
\begin{align*}
    &cos\_sim(\overrightarrow{\textnormal{large}} - \overrightarrow{\textnormal{small}}, \overrightarrow{\textnormal{bear}})
    \vspace{-0.5em}
\end{align*}
\noindent With good word embeddings, larger nouns are expected to have higher projection scores. The validity of the adjective scales from word representations is shown by \citet{kim-de-marneffe-2013-deriving}.

\subsection{Measuring Model Reporting Bias}

We measure the reporting bias of our models by comparing model performance on datasets with different levels of reporting bias and on the \Single{}, \Multi{}, \Any{} groups of the {\dataset} dataset.

We assume that CoDa contains no reporting bias, in which case we can interpret \cref{data-eval} as showing that {\dataset} contains a relatively small amount of it, and Wikipedia contains a relatively large amount. Thus, a larger correlation of model outputs with {\dataset} and a smaller one with Wikipedia would indicate less model reporting bias.

Also, since the \Single{} group subjects are those whose attribute distribution concentrates on a single attribute, these subject-attribute pairs are less likely to be reported in text corpora or even image annotations. Therefore, lower model correlation on the \Single{} group than the \Multi{} and the \Any{} groups would be a sign of model reporting bias.

\section{Results}


The experimental results show that multimodal models outperform text-only models, suggesting their advantage in capturing visual commonsense. However, all models are subject to the influence of reporting bias, as they correlate better with the distributions from Wikipedia than those from CoDa and {\dataset}. Prompt tuning and knowledge distillation substantially enhance model performance, while increasing model size does not.

\subsection{Results with MLM Objective}

\paragraph{Color, Shape, Material}

The resulting model performance for the ``average template'' mode is shown in  \cref{zero_shot_avg_all}. Prompt tuning is done in this mode only. Note that because the top-1 accuracy is taken among all possible classes of each relation, it should be interpreted together with the number of classes (\cref{data-summary}).

We can see from \cref{zero_shot_avg_all} that Oscar does better than BERT in almost all cases. Significant difference between Oscar (base) and BERT (base) is seen in most cases.
Also, after soft prompt tuning, both the Spearman correlation and the accuracy substantially improved. 
Although there is considerable variation of the Spearman correlations, we find consistent improvement per example with both prompt tuning and multimodal pretraining (\cref{sec:linked-plots}).

\cref{zero_shot_avg_all} also shows that knowledge distillation helps improve the performance of BERT in all cases, and the distilled model can sometimes even outperform the teacher model, Oscar.
Moreover, the large version of each model does not always outperform its base counterpart, suggesting that increasing the size of the model does not enhance the model's ability to understand visual commonsense. Instead, training with visually grounded data does.

\begin{figure}[t]
    \centering
    \includegraphics[width=\columnwidth]{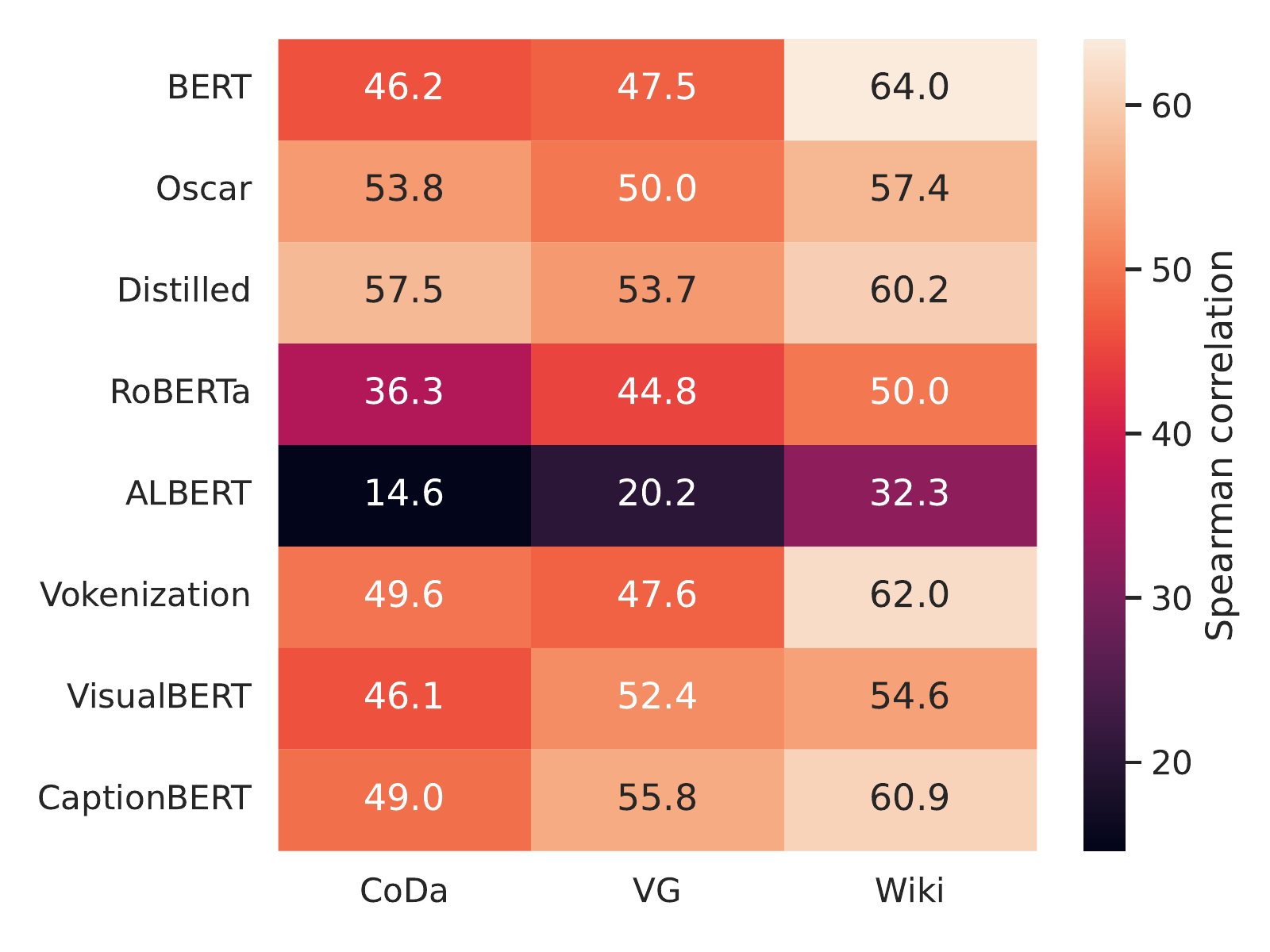}
    \caption{Spearman correlations ($\times 100$) for color, under the ``best template'' case, for base models on CoDa, VG, and Wikipedia. While all models correlate the best with Wikipedia, BERT is the most biased.}
    \label{fig:heatmap}
\vspace{-0.9em}
\end{figure}

\cref{fig:heatmap} illustrates the Spearman correlations of different models with the color distributions from CoDa, {\dataset} and Wikipedia, under the ``best template'' mode.\footnote{\cref{sec:zero-shot} contains further details.} All models correlate moderately with all three datasets, with the highest correlations to Wikipedia, indicating text-based reporting bias in all model types. 
BERT has the largest correlation gap between Wikipedia and CoDa, whereas the visually-grounded models have smaller gaps, indicating less reporting bias in VL models.

\begin{table*}[t!]
\small
\centering
\begin{tabular}{l|lr|lr|lr|l}
\hline
& \multicolumn{2}{c|}{\textbf{Color}} & \multicolumn{2}{c|}{\textbf{Shape}} & \multicolumn{2}{c|}{\textbf{Material}} & \textbf{Co-occur} \\
\cmidrule(lr){2-3}\cmidrule(lr){4-5}\cmidrule(lr){6-7}\cmidrule(lr){8-8}
\textbf{Model} & \textbf{Spearman $\rho$} & \textbf{Acc@1} & \textbf{Spearman $\rho$} & \textbf{Acc@1} & \textbf{Spearman $\rho$} & \textbf{Acc@1} & \textbf{Spearman $\rho$}\\
\hline
BERT$_b$ & 48.0 $\pm$ 21.6 & 51.4 & 53.2 $\pm$ 13.4 & 78.4 & 41.3 $\pm$ 15.6 & 51.1 & 30.2 $\pm$ 11.7 \\
Oscar$_b$ & \textbf{52.5 $\pm$ 20.8} & 63.1 & 54.4 $\pm$ 14.8 & \textbf{80.6} & \textbf{43.2 $\pm$ 14.4} & \textbf{63.0} & 31.2 $\pm$ 12.1\\
CLIP & 51.9 $\pm$ 20.8 & \textbf{63.8} & \textbf{54.5 $\pm$ 13.9} & 79.9 & 42.9 $\pm$ 15.0 & \textbf{63.0} & \textbf{31.3 $\pm$ 11.6}\\
\hline
\end{tabular}
\caption{\label{logistic-reg}
Spearman correlation and top-1 accuracy (both $\times$ 100) with a logistic regression head on model encoder outputs. Oscar and CLIP have comparable performance, both slightly better than BERT.}
\end{table*}

\begin{table*}[t]
\small
\centering
\begin{tabular}{ll|lr|lr|lr}
\hline
&  & \multicolumn{2}{c|}{\textbf{Color}} & \multicolumn{2}{c|}{\textbf{Shape}} & \multicolumn{2}{c}{\textbf{Material}} \\
\cmidrule(lr){3-4}\cmidrule(lr){5-6}\cmidrule(lr){7-8}
\textbf{Group} & \textbf{Model} & \textbf{Spearman $\rho$} & \textbf{Acc@1} & \textbf{Spearman $\rho$} & \textbf{Acc@1} & \textbf{Spearman $\rho$} & \textbf{Acc@1}\\
\hline
\Single{} & BERT$_b$ & 36.8 $\pm$ 19.0 & 54.8 & 48.3 $\pm$ 12.3 & 83.0 & 35.9 $\pm$ 14.3 & 51.6\\
& Oscar$_b$ & 39.9 $\pm$ 15.3 & 60.3 & \textbf{49.3 $\pm$ 11.6} & 87.0 & \textbf{38.5 $\pm$ 12.8} & \textbf{65.1}\\
& CLIP & \textbf{41.0 $\pm$ 15.2} & \textbf{66.3} & 49.2 $\pm$ 14.5 & \textbf{90.0} & 38.1 $\pm$ 12.8 & 64.1\\
\hline
\Multi{} & BERT$_b$ & 49.7 $\pm$ 21.2 & 42.3 & \textbf{65.9 $\pm$ 16.9} & 59.5 & 53.8 $\pm$ 16.2 & 51.3\\
& Oscar$_b$ & \textbf{51.2 $\pm$ 19.9} & 50.6 & 65.2 $\pm$ 17.4 & 64.9 & \textbf{56.2 $\pm$ 13.0} & 53.9\\
& CLIP & 50.5 $\pm$ 21.1 & \textbf{55.4} & 64.6 $\pm$ 18.9 & \textbf{67.6} & 56.2 $\pm$ 14.3 & \textbf{59.2}\\
\hline
\Any{} & BERT$_b$ & 56.5 $\pm$ 19.5 & 46.1 & 100.0 $\pm$ 0 & -- & 58.7 $\pm$ 15.2 & \textbf{35.7}\\
& Oscar$_b$ & \textbf{62.5 $\pm$ 18.9} & \textbf{58.4} & 100.0 $\pm$ 0 & -- & 60.4 $\pm$ 17.1 & \textbf{35.7}\\
& CLIP & 60.3 $\pm$ 18.2 & 55.8 & 100.0 $\pm$ 0 & -- & \textbf{63.5 $\pm$ 20.5} & 21.4\\
\hline
\end{tabular}
\caption{\label{logistic-reg-per-group}
Per-group Spearman correlation and top-1 accuracy (both $\times$ 100) with a logistic regression head on model encoder outputs. Note that the \Any{} group for shape only has one example, so the accuracy is less meaningful and is omitted. All models have higher correlations in the \Multi{} and \Any{} groups than the \Single{} group, which is a sign of reporting bias.}
\vspace{-0.5em}
\end{table*}

\paragraph{Visual Co-occurrence}
\cref{zero_shot_avg_all} also contains the results on visual co-occurrence before and after prompt tuning.
Only the Spearman correlations are reported, because the top-1 accuracy is meaningless due to the large number of possible co-occurring objects with any noun.

Before prompt tuning, BERT has small Spearman correlations, suggesting that it may contain little knowledge about the visual co-occurrence relationship. Oscar demonstrates more such knowledge under the zero-shot setting. After prompt tuning, all model performances improve.

\subsection{Results with Classification Head}

\cref{logistic-reg} shows the results of BERT, CLIP, and Oscar when topped with a classification head. We observe that Oscar and CLIP achieve similar performance and both outperform BERT. Note that, while Visual Genome is part of Oscar's pretraining corpus and one might suspect that that gives it an advantage, CLIP is trained on a large corpus from web search that is unrelated to Visual Genome. Therefore, we can conclude that multimodal models pretrained on both images and text outperform text-only models.

\cref{logistic-reg-per-group} breaks down the results in \cref{logistic-reg} into three subject groups. Oscar and CLIP outperform BERT in almost all cases. The top-1 accuracy is higher for the \Single{} group than for the \Multi{} and \Any{} groups, perhaps because the \Single{} group subjects have only one most likely target attribute, which may be easier to predict. Note that the Spearman correlations for all three models become higher from group \Single{} to \Multi{} to \Any{}. \citet{paik-etal-2021-world} argue that higher correlation for the \Any{} and \Multi{} groups is a sign of model reporting bias, as objects in those two groups are more often reported. Thus, the results here indicate that reporting bias is still present in multimodal models.

\subsection{Results: Size Relation}

\cref{size_table} shows results of the rank partition method (\cref{size-eval}), before and after prompt tuning. Surprisingly, prompt tuning does not help in this case. Moreover, the performance for the ``larger'' templates is higher than that of the ``smaller'' templates, suggesting that the models contain inherent preference towards the ``larger'' templates.

\begin{table}[t!]
\small
\centering
\begin{tabular}{clrr}
\hline
\textbf{Tune} & \textbf{Model} & \textbf{Larger} & \textbf{Smaller} \\
\hline
N & BERT$_b$ & 80.0 & 67.1 \\
 & Oscar$_b$ & 79.5 & 67.7 \\
 & Distilled & \textbf{84.6} & 60.7 \\
 & BERT$_l$ & 80.9 & 66.1 \\
 & Oscar$_l$ & 79.4 & \textbf{70.7} \\
\hline
Y & BERT$_b$ & 69.9 & 55.7 \\
 & Oscar$_b$ & 70.6 & 57.3 \\
 & Distilled & 70.6 & 57.3 \\
 & BERT$_l$ & 70.0 & 55.7 \\
 & Oscar$_l$ & 70.6 & 57.3 \\
\hline
\end{tabular}
\caption{\label{size_table}
Percent correct for size relation, for ``larger'' and ``smaller'' templates, before and after soft prompt tuning. Interestingly, tuning does not help with size.}
\vspace{-0.6em}
\end{table}

\begin{figure}[t]
    \centering
    \includegraphics[width=\columnwidth]{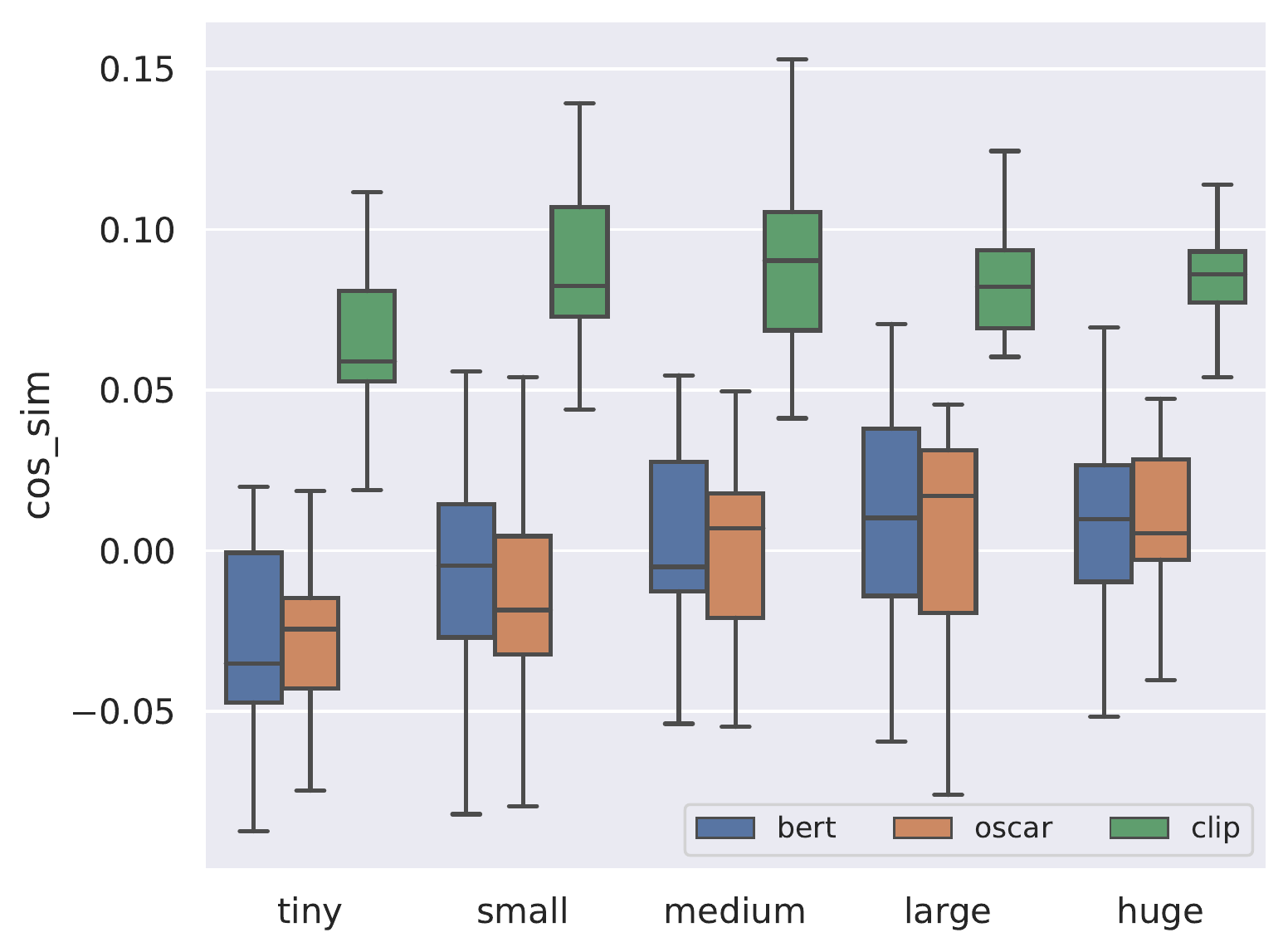}
    \caption{The size projection scores, where the x-axis indicates the object groups. Outliers are omitted. All three models perform reasonably well, as larger objects have higher cosine similarities in general.}
    \label{fig:size_boxplot}
\vspace{-1em}
\end{figure}

\cref{fig:size_boxplot} shows the results of the adjective projection method.\footnote{ \cref{sec:size-per-object} contains per-object plot for BERT vs Oscar.} For BERT and Oscar, we use the average embedding of the subword tokens of the nouns projected onto that of the adjectives ``large'' and ``small''. For CLIP, we take the  textual encoder outputs as the embeddings, resulting in a different score range from that of BERT and Oscar. The results show the following trend: larger objects are projected onto the ``large'' end of the spectrum, although the trend is sometimes broken towards the ``huge'' end. This may be due to the ``huge'' group including nouns such as ``pool'' and ``house'' which can be modified by a relative size indicator ``small''.

\subsection{Analysis and Limitations}

In \cref{zero_shot_avg_all}, the accuracy of BERT for shape is particularly low (only 6.7\%), despite that shape has only 12 classes. We hypothesize that this is due to reporting bias on shape in the text corpora that BERT is trained on. This hypothesis is supported by mining sentences from Wikipedia that contain (noun, attribute) pairs, where we see that the relation shape has fewer number of occurrences than material and color (\cref{sec:wiki-data}).




We also investigate whether the advantage of the visually-grounded models over pure-language models comes from the domain difference between web corpora and image captions, or the presence of actual visual input. Although its teacher is trained with visual inputs, the Distilled model is trained only on captions data and its performance matches that of Oscar, so we hypothesize that grounded training data enhance models' ability to capture visual commonsense. 
The CaptionBERT results supports the hypothesis in favor of domain difference, since it performs better than BERT in both CoDa and VG (\cref{fig:heatmap}). Nevertheless, the visual inputs also have an effect, as Oscar has a higher correlation than CaptionBERT on CoDa.
Thus it seems that both domain and modality affect the ultimate model performance.

Finally, although multimodal models show improvement on the task, sometimes the improvement is not significant and the resulting correlations are still weak. Further work is needed to enhance the visual commonsense abilities of the models and mitigate reporting bias, and our datasets can serve as an evaluation method.


\section{Conclusion}

In this paper, we probe knowledge about visually salient properties from pretrained neural networks. We automatically extract dataset of five visual relations: color, shape, material, size, and co-occurrence, and show that our {\dataset} dataset has a much higher correlation with human perception data for color than data mined from Wikipedia. 
We then apply several probing techniques and discover that visually-supervised models perform better than pure language models, which indicates that they can better capture such visual properties. 
Distilling the knowledge from a visually-supervised model into a pure language model results in comparable performance with the teacher model.

We also observe less reporting bias in both visually-grounded text (VG-mined datasets) than Wikipedia text and visually-grounded models (Oscar, DistilledOscar, VisualBERT, and CLIP) than pure language models. However, visually-grounded models are still subject to the influence of reporting bias, as seen in the per-group analysis, where both types of models perform better for the \Multi{} group than the \Single{} group.

\section*{Acknowledgments}
We would like to thank the reviewers for their comments and suggestions. 
Chenyu Zhang is supported by the Pistritto Research Fellowship.
Elias Stengel-Eskin is supported by an NSF Graduate Research Fellowship. 
Zhuowan Li is supported by NSF 1763705.


\newpage

\bibliography{anthology,custom}
\bibliographystyle{acl_natbib}

\newpage

\appendix
\section{Appendix}
\subsection{List of Objects}
\label{sec:list-objs}

\cref{obj-ls} shows the list of all possible attributes for relations color, shape, and material. \cref{obj-size} shows the list of objects in the five categories of relation size. Visual co-ocurrence has a large number of objects that are not listed here for space reasons.

\begin{table}[ht]
\small
\centering
\begin{tabular}{ll}
\hline
\textbf{Relation} & \textbf{Classes}\\
\hline
Color & \textbf{black}, \textbf{blue} (aqua, azure, cyan, indigo, navy), \\
& \textbf{brown} (khaki, tan), \textbf{gray} (grey), \\
& \textbf{green} (turquoise), \textbf{orange} (amber), \\
& \textbf{pink} (magenta), \textbf{purple} (lavender, violet), \\
& \textbf{red} (burgundy, crimson, maroon, scarlet), \\
& \textbf{silver}, \textbf{white} (beige), \\
& \textbf{yellow} (blond, gold, golden) \\
Shape & \textbf{cross}, \textbf{heart}, \textbf{octagon}, \textbf{oval}, \\
& \textbf{polygon} (heptagon, hexagon, pentagon), \\
& \textbf{rectangle}, \textbf{rhombus} (diamond), \textbf{round} (circle), \\
& \textbf{semicircle}, \textbf{square}, \textbf{star}, \textbf{triangle} \\
Material & \textbf{bronze} (copper), \textbf{ceramic}, \textbf{cloth}, \textbf{concrete}, \\
& \textbf{cotton}, \textbf{denim}, \textbf{glass}, \textbf{gold}, \textbf{iron}, \textbf{jade}, \\
& \textbf{leather}, \textbf{metal}, \textbf{paper}, \textbf{plastic}, \textbf{rubber}, \\
& \textbf{stone} (cobblestone, slate), \textbf{tin} (pewter), \\
& \textbf{wood} (wooden) \\
\hline
\end{tabular}
\caption{\label{obj-ls}
List of all objects for relation color, shape, and material. Inside the parentheses are the attributes that are grouped into the object class.}
\end{table}

\begin{table}[ht]
\small
\centering
\begin{tabular}{ll}
\hline
\textbf{Size} & \textbf{Objects}\\
\hline
Tiny & ant, leaf, earring, candle, lip, ear, eye, \\
& nose, pebble, shrimp, pendant, spoon, dirt, \\
& pill, bee \\
Small & bird, tomato, pizza, purse, bowl, cup, \\
& mug, tape, plate, potato, bottle, faucet, \\
& pot, knob, dish, book, laptop, menu, \\
& flower, pillow, clock, teapot, lobster, duck, \\
& balloon, helmet, hand, face, lemon, microphone, \\
& foot, towel, shoe \\
Medium & human, door, dog, cat, window, lamp, \\
& chair, tire, tv, table, desk, sink, guitar, \\
& bicycle, umbrella, printer, scooter, pumpkin, \\
& monitor, bag, coat, vase, deer, horse, kite \\
Large & elephant, car, tree, suv, pillar, stairway, \\
& bed, minivan, fireplace, bus, boat, cheetah, \\
& wall, balcony, bear, lion \\
Huge & building, airplane, plane, clocktower, tower, earth, \\
& pool, mountain, sky, road, house, hotel, \\
& tank, town, city, dinasour, whale, school \\
\hline
\end{tabular}
\caption{\label{obj-size}
List of objects in five size categories.}
\end{table}

\subsection{Additional Probing}

\begin{table*}[ht]
\small
\centering
\begin{tabular}{l|lr|lr|lr|r}
\hline
& \multicolumn{2}{c|}{\textbf{Color}} & \multicolumn{2}{c|}{\textbf{Shape}} & \multicolumn{2}{c|}{\textbf{Material}} & \textbf{Cooccur}\\
\cmidrule(lr){2-3}\cmidrule(lr){4-5}\cmidrule(lr){6-7}\cmidrule(l){8-8}
\textbf{Model} & \textbf{Spearman $\rho$} & \textbf{Acc@1} & \textbf{Spearman $\rho$} & \textbf{Acc@1} & \textbf{Spearman $\rho$} & \textbf{Acc@1} & \textbf{Spearman $\rho$}\\
\hline
BERT$_b$ & 47.5 $\pm$ 21.6 & 41.8 & 48.2 $\pm$ 12.0 & 64.3 & 41.9 $\pm$ 15.4 & 55.3 & 6.1 $\pm$ 4.0\\
Oscar$_b$ & 50.0 $\pm$ 19.8 & 59.8 & \textbf{52.7 $\pm$ 10.0} & 89.3 & \textbf{46.5 $\pm$ 13.7} & \textbf{74.6} & 10.1 $\pm$ 7.2\\
Distilled & 53.7 $\pm$ 21.3 & 57.7 & 51.4 $\pm$ 11.1 & 74.3 & 46.0 $\pm$ 13.6 & \textbf{74.6} & 10.4 $\pm$ 7.8\\
RoBERTa$_b$ & 44.8 $\pm$ 19.8 & 41.6 & 45.4 $\pm$ 12.4 & 69.3 & 33.0 $\pm$ 15.5 & 39.1 & 1.1 $\pm$ 1.4\\
ALBERT$_b$ & 20.2 $\pm$ 24.8 & 13.4 & 29.8 $\pm$ 15.7 & 13.6 & 25.0 $\pm$ 17.9 & 27.8 & 6.6 $\pm$ 5.1\\
Vokenization & 47.6 $\pm$ 20.9 & 51.6 & 49.8 $\pm$ 13.1 & 72.9 & 39.4 $\pm$ 16.0 & 52.5 & 6.0 $\pm$ 3.7\\
VisualBERT & 52.4 $\pm$ 19.8 & 65.3 & 48.7 $\pm$ 12.9 & 66.4 & 43.4 $\pm$ 15.5 & 59.5 & \textbf{10.7 $\pm$ 8.1}\\
CaptionBERT & \textbf{55.8 $\pm$ 20.6} & \textbf{70.0} & 51.3 $\pm$ 11.8 & \textbf{91.4} & 42.6 $\pm$ 15.4 & 54.6 & 10.2 $\pm$ 7.5\\
\hline
\end{tabular}
\caption{\label{zero_shot_best_all}
Spearman correlation and top-1 accuracy (both $\times$ 100) of zero shot probing. This is the ``best template'' case discussed in \cref{models}.}
\end{table*}

\paragraph{Best template mode}
\label{sec:zero-shot}
\cref{zero_shot_best_all} contains zero-shot results under the ``best template'' mode, for BERT (base), Oscar (base), BERT distilled from Oscar, RoBERTa (base), ALBERT (base), Vokenization, and VisualBERT (base). These results demonstrate similar trends as the ones in the ``average template'' mode.

\paragraph{Per-object analysis}

\begin{figure}[ht]
    \centering
    \includegraphics[width=\columnwidth]{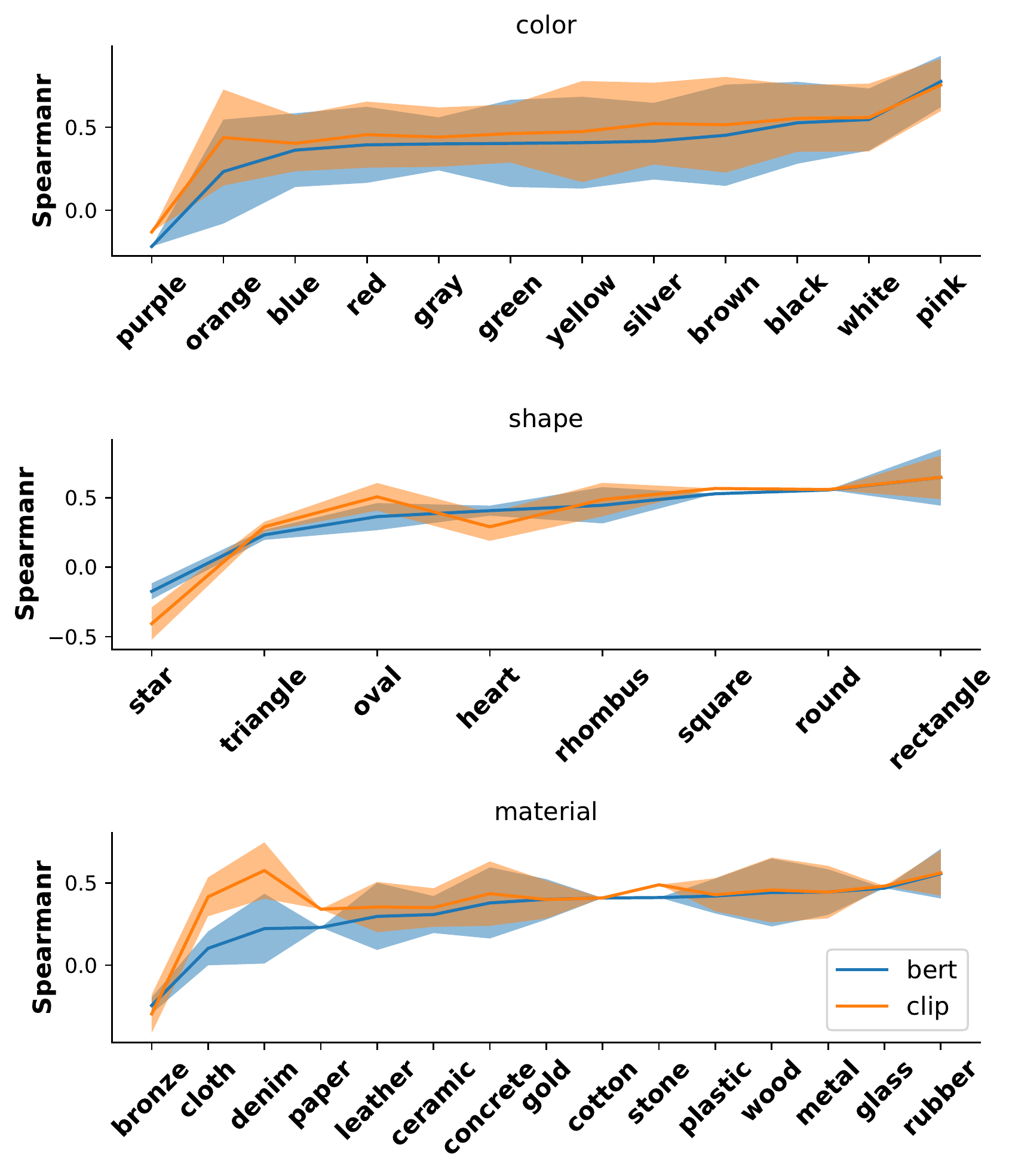}
    \caption{Spearman correlation per object class for BERT and CLIP with the logistic regression head, for color, shape, and material. The error margins are the standard deviations.}
    \label{fig:logsitic_reg_fig}
\end{figure}

\cref{fig:logsitic_reg_fig} illustrates the fine-grained Spearman correlation $\pm$ standard deviation per object group for BERT and CLIP.

\paragraph{Size per-object}
\label{sec:size-per-object}

\begin{figure}[h]
    \includegraphics[width=\columnwidth]{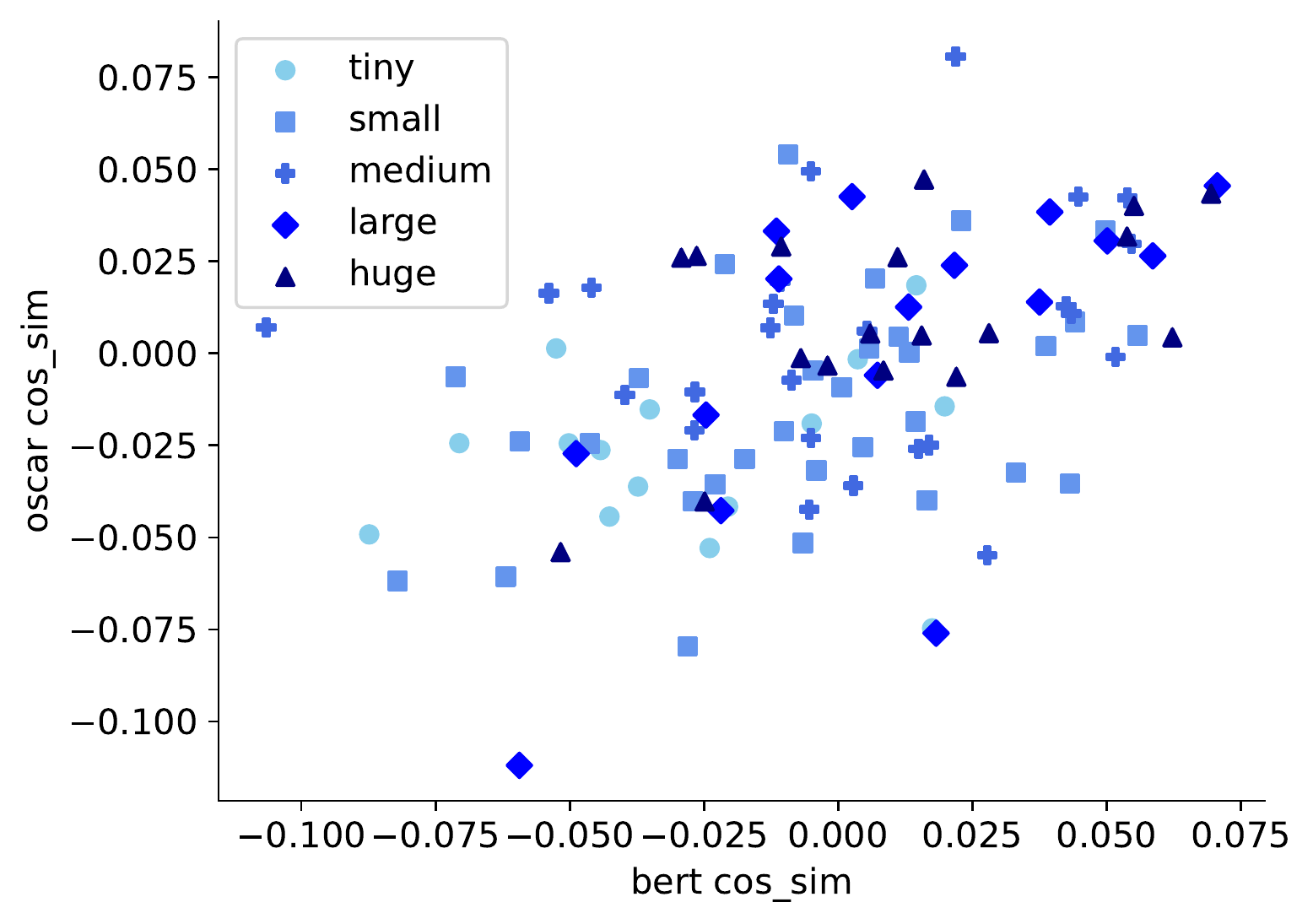}
    \caption{The size projection scores from BERT and Oscar, where each point is one object. Cosine similarities are correlated between Oscar and BERT.}
    \label{fig:size_scatterplot}
\end{figure}

\cref{fig:size_scatterplot} shows how the per-object projection scores on the size spectrum from BERT and Oscar are correlated.

\paragraph{Per-Subject Comparison}
\label{sec:linked-plots}

\begin{figure*}[ht]
    \centering
    \includegraphics[width=\textwidth]{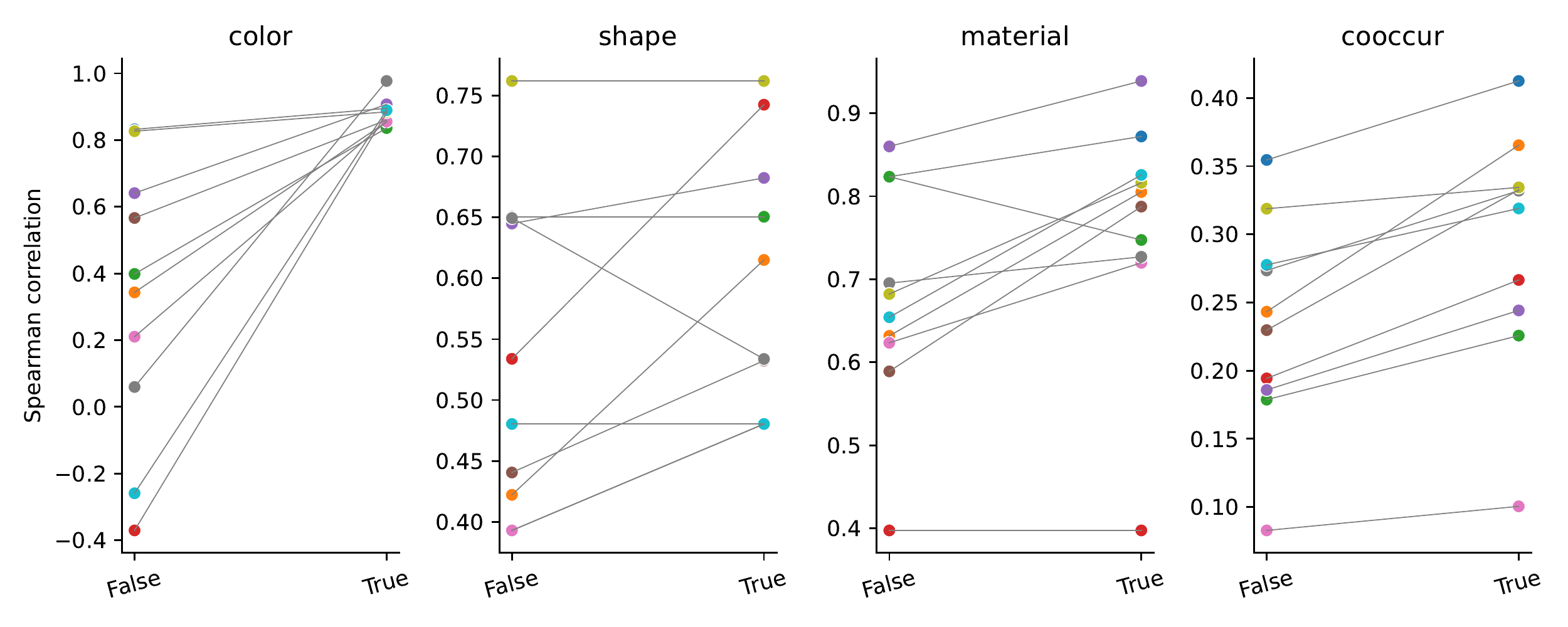}
    \caption{Spearman correlation of 10 subjects for each relation type before and after soft prompt tuning, with Oscar (base). Almost all individual subject has increased correlation after prompt tuning, except in relation shape.}
    \label{fig:linked_tune_fig}

    \includegraphics[width=\textwidth]{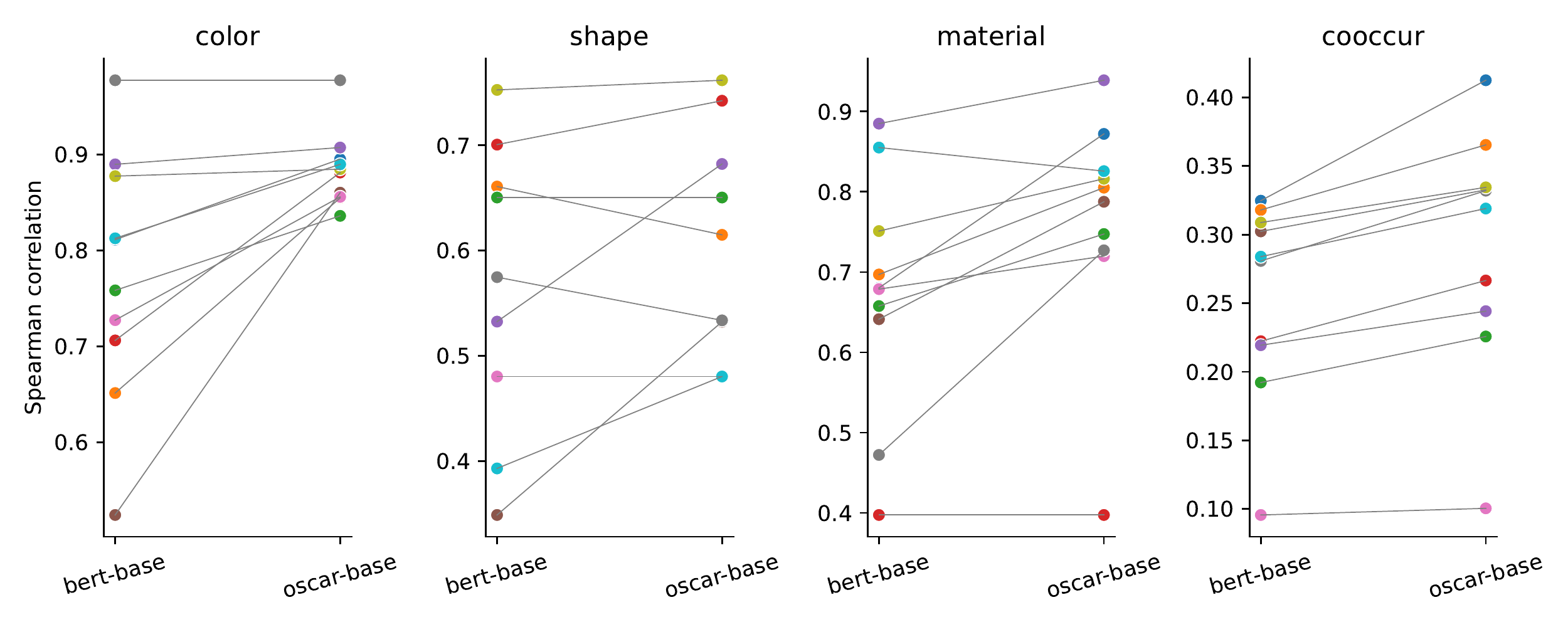}
    \caption{Spearman correlation of 10 subjects for each relation type with BERT (base) and Oscar (base), after soft prompt tuning. Almost all individual subject has higher correlation with Oscar than with BERT, except in relation shape.}
    \label{fig:linked_model_fig}
\end{figure*}

\cref{fig:linked_tune_fig} and \cref{fig:linked_model_fig} show how the Spearman correlations of 10 individual subjects improve after soft prompt tuning and after multimodal pretraining. Consistent improvement can be seen in color, material, and cooccurrence. 
Although we report average Spearman correlations in \cref{zero_shot_avg_all} and there are large standard deviations, here we show that when improvement is observed collectively, it is also consistent across subjects. With shape, the improvement is less obvious (45.9 to 50.4 for prompt tuning and 49.2 to 50.4 for multimodal pretraining).

\subsection{Error Analysis}

\paragraph{Data}
\label{sec:data-analysis}

\begin{figure}[ht]
    \centering
    \includegraphics[width=\columnwidth]{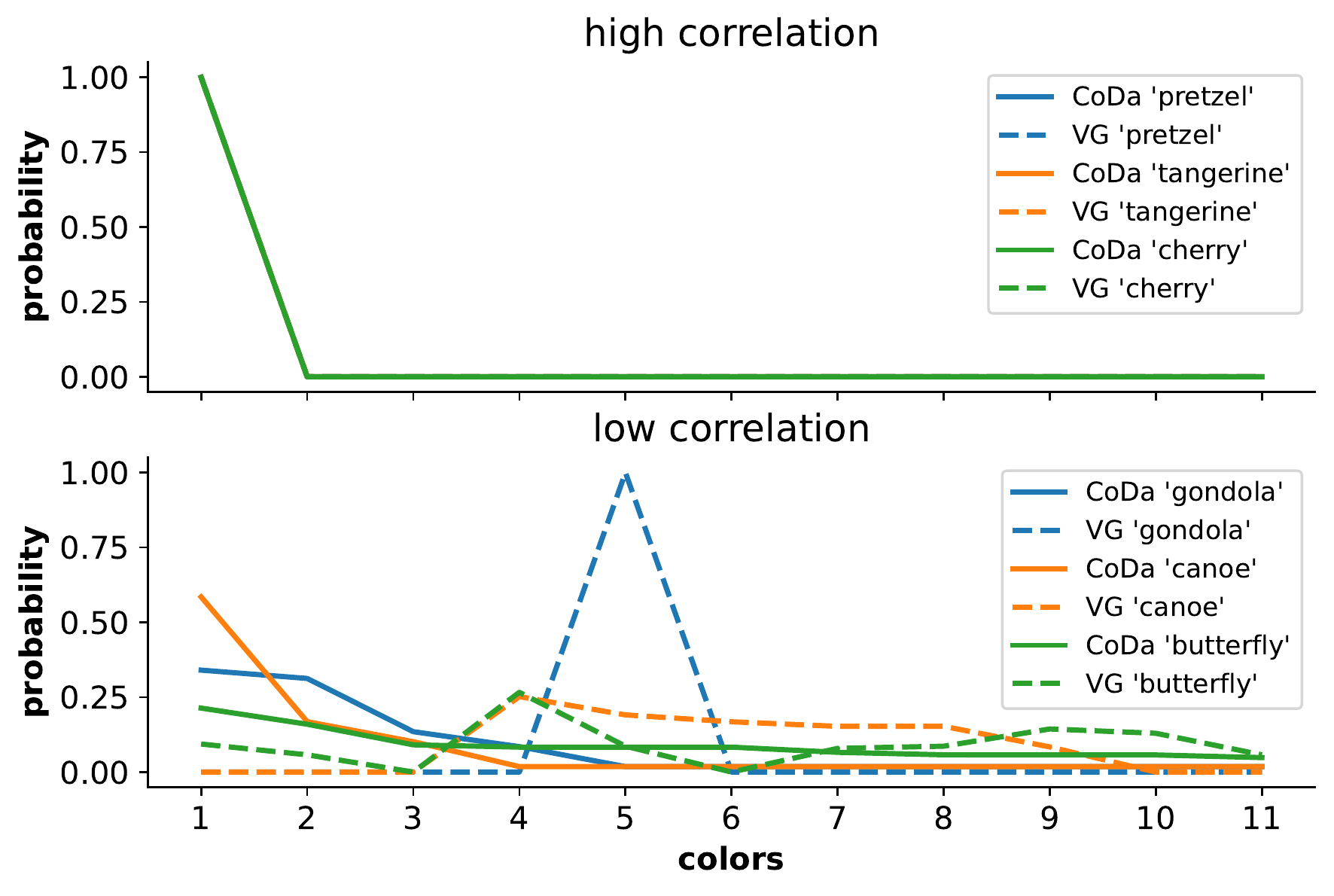}
    \caption{VG vs. CoDa distribution of 3 subjects with the lowest and highest correlation, ordered by probability of colors in CoDa.}
    \label{fig:eval_data_all}

    \includegraphics[width=\columnwidth]{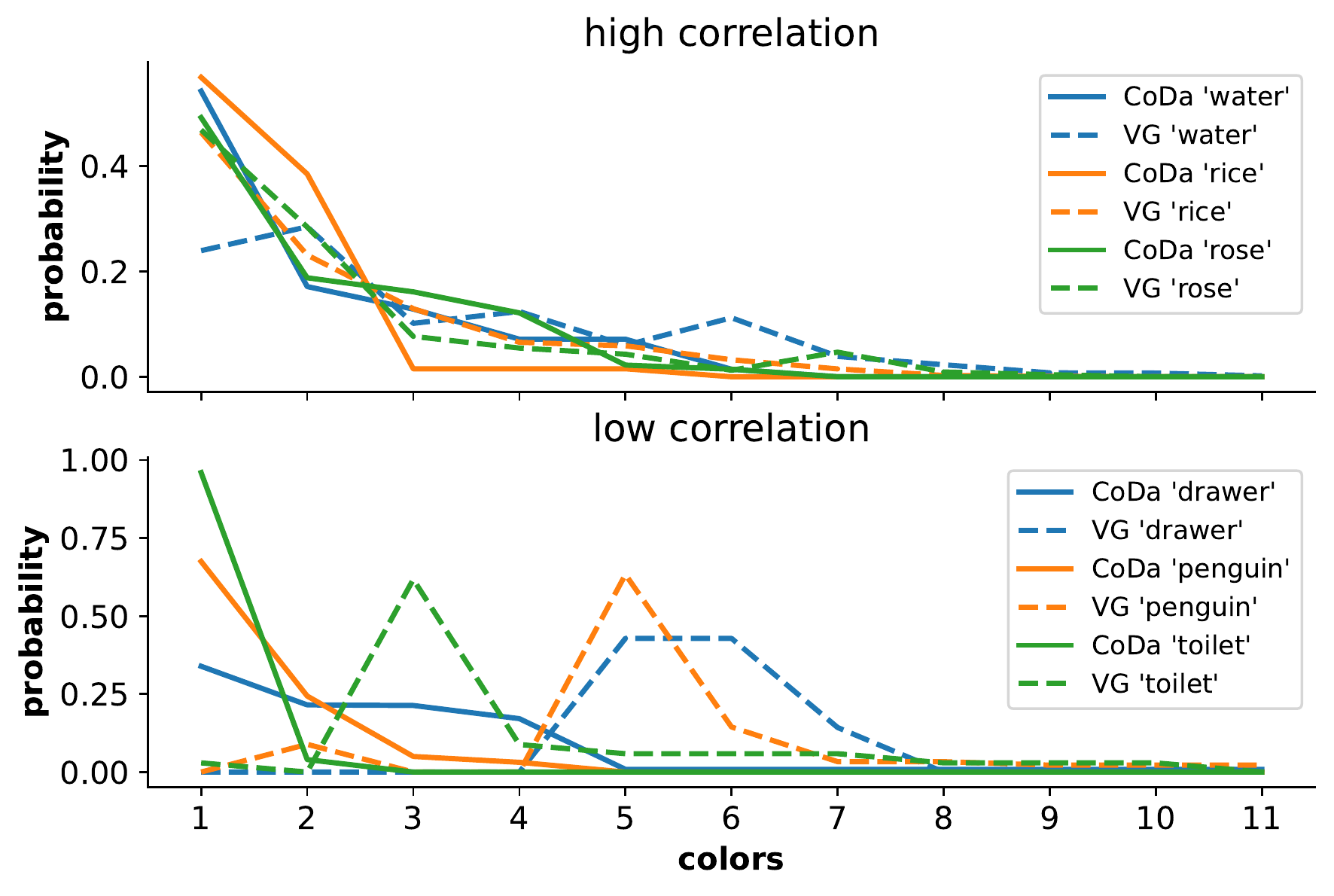}
    \caption{Wikipedia vs. CoDa distribution of 3 subjects with the lowest and highest correlation, ordered by probability of colors in CoDa.}
    \label{fig:eval_data_wiki_all}
\end{figure}

The three subjects with the highest and lowest Spearman correlation are shown in \cref{fig:eval_data_all} and \cref{fig:eval_data_wiki_all}.

\paragraph{Wikipedia}
\label{sec:wiki-data}
\cref{wiki-count} shows the number of (noun, attribute) pairs of the three relation types in Wikipedia. Shape has fewer occurrences than material and color.

\begin{table}[h]
\small
\centering
\begin{tabular}{lrrr}
\hline
 & \textbf{Color} & \textbf{Shape} & \textbf{Material} \\
\hline
Total & 331480 & 195921 & 307879 \\
Avg 12 & 27623.3 & 16326.8 & 24634.7 \\
\hline
\end{tabular}
\caption{\label{wiki-count}
First row is the total number of occurrences of (noun, attribute) pairs for relations shape, material, and color in Wikipedia. Second row is the average number of occurrences across the top 12 attributes for each relation. Shape has the fewest number of occurrences.}
\end{table}

\paragraph{Model}

\begin{table*}[ht]
\small
\centering
\begin{tabular}{l|ll|ll}
\hline
& \multicolumn{2}{c|}{\textbf{High Corr Subjs}} & \multicolumn{2}{c}{\textbf{Low Corr Subjs}}\\
\cmidrule(lr){2-3}\cmidrule(lr){4-5}
\textbf{Relation} & \textbf{BERT$_b$} & \textbf{Oscar$_b$} & \textbf{BERT$_b$} & \textbf{Oscar$_b$}\\
\hline
Color & lace, jacket, design & balloon, jacket, apple & flush, water faucet, muffler & hinge, leg, slack\\
Shape & mirror, vase, container & chair, pizza, vase & connector, log, knot & banana, toast, phone\\
Material & wall, tray, board & fence, wall, shelf & sheep, fabric, patch & elephant, rug, patch\\
\hline
\end{tabular}
\caption{\label{error-average}
Three subjects each with high and low correlations for relations color, shape, and material.}
\end{table*}

\cref{error-average} shows the errors made by BERT and Oscar in the ``average template'' mode before prompt tuning. Overall, subjects with low correlation are those that are less often reported in Visual Genome as well as in textual data.

\subsection{Resources}
\label{resources}

\paragraph{BERT, RoBERTa, ALBERT}
We use the Huggingface implementations of BERT, RoBERTa, and ALBERT.

\paragraph{Oscar}
See the GitHub repository for the code and pretrained Oscar: \url{https://github.com/microsoft/Oscar}.

\paragraph{CLIP}
We use the CLIP model released by OpenAI: \url{https://github.com/openai/CLIP}.

\paragraph{Vokenization}
See the GitHub repository for the pretrained model: \url{https://github.com/airsplay/vokenization}.

\end{document}